\documentclass[5p, authoryear, 10pt]{elsarticle}

\usepackage[english]{babel}
\usepackage[utf8x]{inputenc}
\usepackage[T1]{fontenc}

\usepackage{amsmath}
\usepackage{graphicx}
\usepackage{url}            
\usepackage{booktabs}       
\usepackage{amsfonts}       
\usepackage{nicefrac}       
\usepackage{microtype}      
\usepackage[authoryear,round]{natbib}

\usepackage{hyperref}
\hypersetup{pdfpagemode=UseNone, pdfstartview=FitH,
  colorlinks=true,urlcolor=black,linkcolor=black,citecolor=black}

\usepackage[colorinlistoftodos,prependcaption]{todonotes}
\usepackage{regexpatch}
\makeatletter
\xpatchcmd{\@todo}{\setkeys{todonotes}{#1}}{\setkeys{todonotes}{inline,#1}}{}{}
\makeatother

\journal{Neuroimage}

\begin{document}
\begin{frontmatter}

\title{Adaptive neural network classifier for decoding MEG signals}
\author[nbe]{Ivan Zubarev\corref{cor}}
\author[nbe]{Rasmus Zetter}
\author[nbe]{Hanna-Leena Halme}
\author[nbe,ani]{Lauri Parkkonen}

\address[nbe]{Department of Neuroscience and Biomedical Engineering, Aalto University School of Science, FI-00076 Aalto, Finland}
\address[ani]{Aalto NeuroImaging, Aalto University, FI-00076 Aalto, Finland}

\cortext[cor]{Corresponding author \href{mailto:ivan.zubarev@aalto.fi}{ivan.zubarev@aalto.fi}}

\begin{abstract}
Convolutional Neural Networks (CNN) outperform traditional classification methods in many domains. Recently these methods have gained attention in neuroscience and particularly in brain–computer-interface (BCI) community. Here, we introduce a CNN optimized for classification of brain states from magnetoencephalographic (MEG) measurements. Our CNN design is based on a generative model of the electromagnetic (EEG and MEG) brain signals and is readily interpretable in neurophysiological terms. We show here that the proposed network is able to decode event-related responses as well as modulations of oscillatory brain activity and that it outperforms more complex neural networks and traditional classifiers used in the field. Importantly, the model is robust to inter-individual differences and can successfully generalize to new subjects in offline and online classification.
\end{abstract}

\begin{keyword}
convolutional neural network \sep magnetoencephalography \sep brain--computer interface

\end{keyword}

\end{frontmatter}

\section{Introduction}

Deep Neural Networks have revolutionized many domains such as image recognition and natural language processing. To date, their application in the analysis of electro- and magnetoencephalographic (EEG and MEG) data has been limited by several domain-specific factors.

First of all, electromagnetic brain signals are characterized by very low signal-to-noise ratio (SNR). Here, the term "noise" is understood widely and includes external interference, physiological (e.g. cardiac or oculomotor) artifacts as well as background brain activity unrelated to the studied phenomena. SNR in single-trial EEG and MEG measurements is typically assumed to be $< 1$ for evoked responses and $\mathbf{\approx}1$ for oscillatory activity, which puts these data to stark contrast with those in traditional applications of deep learning. Typical EEG/MEG analysis employs a wide range of techniques to increase the SNR, e.g. spatial and temporal filtering, averaging a large number of observations, source-separation algorithms, and other complex feature extraction methods (e.g. wavelet transform). Thus, efficient noise suppression is required for high-accuracy classification of EEG/MEG signals.

Second, these data have a complex, high-dimensional spatiotemporal structure. Modern EEG and MEG systems comprise several hundreds of sensors capable of sampling brain activity with sub-millisecond temporal resolution. In case of MEG, these sensors may also measure different components of the neuromagnetic field. On one hand, this multitude of data points enables sophisticated experimental designs and analysis methods to extract finer details of brain function. On the other hand, manual analysis and interpretation of these data becomes increasingly complex and time-consuming. Machine-learning algorithms can be of great help in such tasks but the mere classification result is often not sufficient; ideally, the experimenter should understand why the algorithm is able classify the data, i.e., the learned model should be interpretable in neurophysiological terms. A model able to reliably identify those neural sources that contribute to the discrimination between given experimental conditions could enable efficient exploitative analysis of these complex data sets and ultimately allow more complex experimental designs.

Finally, sample sizes in EEG/MEG data sets are typically too small for deep-learning models. One reason for this is the high cost and time limit of collecting these data. Running an experiment on a single human subject for many hours to collect a large enough data set is often not feasible. On the other hand, pooling data from many subjects is a promising strategy to overcome this limitation, but it requires the classifier to be robust to high inter-individual variability stemming from differences in cortical anatomy, physiological state etc.

Taken together, these factors may easily lead to over-fitting (especially in more complex models) and poor interpretability of the models. To address these challenges, we propose a Convolutional Neural Network (CNN) whose architecture is based on a generative model of non-invasive electromagnetic measurements of the brain activity \citep{Daunizeau2007AMEG}. 

This network utilizes spatiotemporal structure in the MEG data to extract informative components of MEG signal from the noisy observations. Since the model structure reflects our understanding of the data generation process, the extracted components can be interpreted in terms of the underlying neural activity. Specifically, this model assumes that the MEG measurements are generated by a linear (spatial) mixture of a limited number of latent sources, which evolve non-linearly over time. Each of these sources is characterized by a spatial distribution and a time course, which are relatively stable within each individual but may vary across individuals. For example, the spatial topography of a sensory evoked response may vary across subjects due to small differences in cortical anatomy but its latency can be relatively constant. Conversely, the phase of an event-related (induced) oscillatory response may vary considerably across trials while its spectral content and spatial distribution remain the same. 

We demonstrate that utilizing a generative model makes the algorithm robust to inter-individual differences in spatial, temporal and spectral properties of the signal. Critically, such across-subject generalization makes it possible to train the model on pooled data from multiple subjects and successfully apply it to new subjects. Here, we applied this neural network to classify evoked responses to visual, auditory and somatosensory stimuli (5 classes; Experiment 1) and induced responses to hand motor imagery (3 classes; Experiment 2). Finally, we ran the algorithm in real time and tested it in a brain–computer interface (BCI) (2 classes; Experiment 3).

\section{Methods}

\subsection{Generative latent state-space model}
Magnetoencephalography (MEG) is a non-invasive, time-resolved technique for measuring electric brain activity through the magnetic field it generates \citep{Hamalainen1993}. The MEG signal is complementary to that of electroencephalography (EEG), in which the potential distribution caused by electric brain activity is measured using electrodes placed on the scalp. MEG is considered to have higher spatial resolution than EEG, as the EEG signal is distorted by the heterogeneous conductivity profile of head tissues to a much larger extent than the MEG signal \citep[see e.g.][]{Baillet2017}.

MEG data typically include 1) stereotyped evoked responses (event-related fields; ERF) that are phase-locked to specific sensory, cognitive or motor events, and 2) induced modulations of ongoing oscillatory brain activity that is not phase-locked to external events. MEG measurements are typically contaminated by noise and interference originating from external sources as well as by ongoing unrelated brain activity. Single-trial ERFs typically have a signal-to-noise ratio (SNR) $\sim$ 1. 

An MEG measurement can be represented by an $n \times t$ data matrix $\mathbf{X}$ containing measurements from $n$ sensors (magnetometers or gradiometers; typ. 200–300) at $t$ time points sampled at a high temporal frequency (typ. $\sim$1000 Hz). These data have a complex spatiotemporal structure because an activation of a single neural source is picked up by several sensors at different spatial locations and these signals exhibit temporal correlations. Thus, simultaneously active neural sources result in a high degree of linear spatial mixing as well as non-linear temporal dependencies in the measured data. Fortunately, dense spatial and temporal sampling allow efficient source-separation by utilizing local spatiotemporal correlations\citep{Cardoso1998BlindPrinciples}. We argue that an effective approach towards decoding brain states should take into account these properties of the signal.

The proposed network architecture is broadly based on an extension of a model describing the generation of MEG signal \citep{Daunizeau2007AMEG}. The model is motivated by the assumption that a single event-related MEG observation $\mathbf{X} \in$ $\mathbb{R}^{n \times t}$ is generated by a mixture of $k$ latent sources $s$ such that at each time point $t$

\begin{equation}
\mathbf{x}_t = \mathbf{C s}_t + \mathbf{\epsilon}
\label{eq:general_model}
\end{equation}
where $\mathbf{C}$ is an $n \times k$ matrix describing the spatial mixing of the $k$ underlying latent sources whose time courses $\mathbf{s}$ are the rows of a matrix $\mathbf{S}$ and $\epsilon$ is additive Gaussian white observation noise. The number of such latent sources is small relative to the number of MEG channels (i.e., $k \ll n$). 
These $k$ sources evolve in time in a structured way and may or may not be statistically dependent. 
Importantly, we do not restrict the definition of a latent source to neural activity as some of these sources may correspond to e.g. ocular or muscular artifacts. In the simplest case, the mapping from $\mathbf{S}$ to $\mathbf{X}$ is linear, and non-linearities in temporal dynamics of $\mathbf{S}$ can be locally approximated by a linear autoregressive (AR) model $\mathbf{A}$ of order $L$ with innovation noise $\mathbf{\omega}$. Given the fast temporal sampling of MEG, such local linearity in the temporal domain is a reasonable assumption. Thus,

\begin{equation}
\mathbf{s}_{t} =\sum_{l=1}^{L} \mathbf{A}_l \mathbf{s}_{t-l}
\label{eq:time_evolution}
\end{equation}

Assuming no interaction between the latent sources, $\mathbf{A}_l$ becomes a diagonal submatrix, where the $k$-th diagonal element of each of the $L$ submatrices form a $L-$th order univariate AR model of the temporal dynamics of the $k$-th source. The coefficients of these AR models contain information about spectral properties of the sources. Furthermore, if $\mathbf{A_l}$ is a full submatrix, its off-diagonal elements model the interactions between the latent sources, leading to $\mathbf{A}$ being a full $L$-th order Vector-Autoregressive (VAR) model of $k$ interacting sources. 

\subsection{Network architecture}
The proposed classifier incorporates the assumptions of the generative model described above into the discriminative neural network model. The first and the second layers of the network learn spatial and temporal filters, which extract a compact representation of MEG signal features contributing to the discrimination between the classes. These features make use of spatial and temporal correlations in the data to suppress noise and to obtain sufficient separation between the simultaneously active neural sources. 
The $l_1$-regularized output layer assigns non-zero weights only to features that are informative for each class of the stimuli. Finally, the spatial and temporal filters are further optimized by back-propagating errors from the output-layer nodes with non-zero weights.

\paragraph{Input layer: Spatial de-mixing}
The linear input layer trains a set of spatial filters $\mathbf{W}$ with each column $\mathbf{w}_k$ extracting a timecourse of ${k}$-th latent source. These filters are related to the spatial activation patterns $\mathbf{C}$ of the latent sources in the generative model (Eq. \ref{eq:general_model}) via

\begin{equation}
\mathbf{W}^T \mathbf{x}_t = \mathbf{\hat{s}_t}
\label{eq:filter_pattern}
\end{equation}

\begin{equation}
\mathbf{C} = \mathbf{\Sigma}_x \mathbf{W} \mathbf{\Sigma}^{-1}_{\hat{s}}
\label{eq:filter_pattern_theorem}
\end{equation}
where $\mathbf{\Sigma}_x$ is the spatial data covariance and $\mathbf{\Sigma}^{-1}_{\hat{s}}$ is the precision matrix of the latent time courses\citep{Haufe2014OnNeuroimaging}. 

\begin{figure*}[h]
\centering
\includegraphics[width=0.85\linewidth]{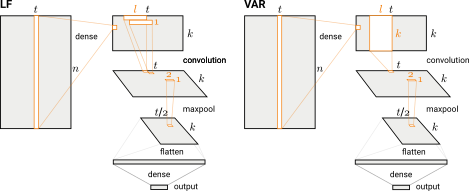}
\caption{Architecture of the two variants of the CNN.}
\label{fig:network_architecture}
\end{figure*}

The input layer can be viewed as a linear projection layer performing dimensionality reduction in the spatial domain. The weights of the input layer are applied to each time point in the MEG epoch by computing dot product between $\mathbf{W}$ and the whole MEG epoch $\mathbf{X}$. 
This layer implements a set of spatial filters that project the channel data onto a $k$-dimensional subspace. This layer has several functions: (1) it obtains a lower-dimensional and spatially-decorrelated representation of the signal time courses, (2) it learns and projects out irrelevant activity such as physiological artifacts, and (3) it provides an interpretable linear mapping from model weights to the channels in the original signal space. These weights can be used to extract spatial activation patterns \citep{Haufe2014OnNeuroimaging}, or topographic maps, corresponding to neural sources informing the classifiers.
Alternatively, this layer can be viewed as a spatial convolution layer with 'valid' padding applied to all channels at each time-point. Here, we prefer to refer to it as a linear projection layer for ease of interpretation. 
Linear projections similar those for the function (2) above are typically used to suppress ocular and cardiac artifacts in MEG data. However, in contrast to e.g. independent component analysis (ICA), the projection basis is defined by back-propagation without introducing an explicit assumption of statistical independence.

\paragraph{Temporal convolution layer: Extraction of activation dynamics} This layer operates on the time courses of the latent sources (corresponding to the rows of $\mathbf{S}$) and implements a filter, which extracts a temporal activation pattern of the informative neural event (e.g. a peak of an evoked response).

We used two variants of this layer. The simpler one (LF-CNN) applies separate 1-dimensional convolution filters of the $l$-th order to the time courses of the $k$ spatial components produced by the input layer. The model assumes that these time courses do not interact and that they have unique spectral fingerprints. This layer variant can be viewed as applying linear finite-impulse-response filters (hence LF) that specifically capture the fingerprint of each spatial component.

The more complex variant allows estimating the interactions between the spatial components and can be viewed as a vector autoregressive model (VAR-CNN) of the component time courses. This structure is implemented by applying $k$ spatiotemporal convolution kernels of shape $l \times k$ and it results in a larger set of trainable parameters ($l k^2$) in this layer. Thus, each spatial pattern from the input layer has a corresponding impulse-response function learned by the temporal convolution layer. These functions extract the frequency bands specific for each component.

For both variants of this layer, the convolution is followed by a non-linearity using rectified linear units ($ReLU$) and a max-pooling layer with a pooling factor of 2 and a stride of 2 applied to the time dimension. Temporal max-pooling provides robustness against variation in the latency of the informative responses across subjects. 

\paragraph{Output layer: Imposing sparsity} The mapping from the temporal convolution layer to the output is provided by a single, fully-connected layer followed by a soft-max normalization. Sparsity is imposed on the weights of the output layer using $l_1$-norm regularization suppressing most of the activity that is unrelated to the classification. Exploring the non-zero weights in the output layer thus allows to identify the temporal and spatial patterns which contribute to the discrimination. 

\subsection{Model inspection and parameter interpretation}

Interpretation of the model parameters in terms of the underlying neural activity is a desirable property. Our network design is based on a generative model of MEG signal to allow such interpretation.

To identify spatial and temporal features that contribute to assignment of a given sample to a particular class, we identified the nodes of the final classification layer containing the maximum positive weights (contributions) to each particular class. 
Since the input to the final layer (before flattening) has the dimensions corresponding to:
\begin{enumerate}
    \item number of latent components
    \item number of (pooled) time points
    \item number of classes 
\end{enumerate}
 we can identify the index of the latent component (or several components) that has a maximum contribution to this class as well as the (approximate) timepoint corresponding to the maximum activation of this component. These indices are then used to extract the corresponding spatial filter from the input layer and temporal convolutional filter from the hidden layer. Thus, a single feature in the output layer maps to the information about spatial pattern, approximate latency and spectral properties of the latent component.

\paragraph{Spatial activation patterns and source estimates}
Assuming statistical independence between the latent components one can extract the corresponding activation pattern by multiplying the spatial filter by the spatial covariance matrix of the data \citep{Haufe2014OnNeuroimaging}. To relax the independence assumption further, one can also consider multiplying the resulting pattern by the precision matrix of the latent time courses that can be obtained by inverting the covariance matrix of the data, projected onto the latent component space with the spatial filters extracted during the previous step. In this study this latter step was omitted, because it is unclear whether the spatial filters optimized to extract only the  informative parts of the latent time course could be used to estimate the latent precision matrix without distortion (Supplementary Figure 1).

The obtained sensor-level spatial pattern can further be mapped onto the individual source space using standard source estimation algorithms (Figure \ref{fig:interpretability}). To demonstrate this mapping we trained an LF-CNN model on pooled data from Experiment 1 and updated it with a single training epoch performed on the data of a single held-out subject. For each of the 5 classes of stimuli, we extracted the model parameters corresponding to spatial filters of the single most informative component using the procedure described above. These patterns were compared to components that had minimal or no contribution to any of the classes as defined by the absolute sum of the corresponding weights in the output layer (Figure \ref{fig:non_informative}).
We then estimated the neural sources of each informative component by computing dynamic statistical parametric maps (dSPM) \citep{Dale2000} for sources constrained onto the individual cortical surface and with orientations favoring the direction perpendicular to the local cortical surface (loose orientation constraint 0.2) as implemented in the MNE-Python software package \citep{Gramfort2013}. 

We then conducted a standard evoked-response analysis by averaging the MEG responses within each class of stimuli. We compared the features extracted from our model to the most prominent components of the corresponding averaged evoked responses in terms of their latency and spatial distribution and the corresponding source estimates. Source estimation of the averaged evoked responses utilized noise covariance estimated from a 300-ms pre-stimulus baseline whereas identity covariance was used for activation patterns.

\paragraph{Investigating spectral properties of extracted latent components}
Since each temporal convolution filter in the hidden layer of the LF-CNN represents a univariate autoregressive model of the temporal dynamics of $k$-th latent source, we can use the coefficients of this filter to obtain an estimate of the power spectral density properties of the time course of this latent source. To test whether this approach can be used to identify oscillatory activity that is informative for classification we extracted the spatial and spectral filters from the model trained on the data from Experiment 2. Because we expected informative activity to be extended over the whole 1.5 s time window, since event-related desynchronization associated with motor imagery is not phase-locked to the stimulus onset, we used a different approach to identify the most informative latent component compared to the Experiment 1. Instead of identifying a single feature in the output layer, corresponding to highest activation at a single time-point, we took the index of spatial component that had the largest sum of weights over all time-points. We extracted spatial patterns corresponding to two most informative latent components for each class. We also estimated the frequency content of the component's time course by computing power spectral density using the weights of the corresponding temporal convolutional filters.

\subsection{Implementation and training}
\paragraph{Design choices and hyperparameters}
The neural network was implemented using the Tensorflow library \citep{Abadi2016}. The code is publicly available at \url{https://version.aalto.fi/gitlab/zubarei1/aalto-megnet}. Model development and hyperparameter tuning were performed on the data of a single randomly-picked subject from Dataset 1. The model was then applied to across-subject classification and other experiments as is. Table \ref{tab:hyperpars} summarizes the tested values of tunable hyperparameters.

\begin{table*}[htbp]
  \centering
  
  \caption{Tested and optimal hyperparameter values.}
  
    \begin{tabular}{lrr}
    \toprule
     Parameter & Tested & Optimal\\\midrule
    Number of latent sources & $16, 32, 64$ & \multicolumn{1}{r}{$32$} \\
    Temporal filter length & $3,5,7,9,11$ & \multicolumn{1}{r}{$7$}\\
    Learning rate & $1\cdot10^{-3}$, $3\cdot 10^{-4}$, $1\cdot 10^{-4}$ & \multicolumn{1}{r}{$3\cdot10^{-4}$}\\
    $l_{1}$-penalty & $1\cdot10^{-3}$, $3\cdot 10^{-4}$, $1\cdot 10^{-4}$ & \multicolumn{1}{r}{$3\cdot10^{-4}$} \\
    
    Pooling & $max$ & \multicolumn{1}{r}{$max$} \\
    Pooling factor & $2,3,5$ & \multicolumn{1}{r}{$2$} \\
    Drop-out coefficient & $0.25, 0.50, 0.75, 0.90$ & \multicolumn{1}{r}{$0.5$0} \\
    Input layer link function & $identity, ReLU, ELU, tanh$ & $identity$ \\
    Hidden layer link function & $identity, ReLU, ELU, tanh $& $ReLU$ \\
    Output nonlinearity & $sigmoid, softmax$ & $softmax$ \\
    Number of dense hidden layers & $1,2$ & $1$\\
    \bottomrule
    \end{tabular}%
    
  \label{tab:hyperpars}%
\end{table*}%

\paragraph{Initialization and training}
The initial values of the weight matrices were drawn from a uniform distribution following the procedure introduced by \citet{He2015DelvingClassification}.

We initialized the bias variables to a constant value of 0.1. 
We used the Adam optimization algorithm with a batch size of 100 and learning rate of $3.0 \cdot 10^{-4}$ to optimize multinomial cross-entropy between the model predictions and true labels. Higher learning rates were also used but they did not improve performance. 
We used an early-stopping strategy to prevent over-fitting; for every 1000 iterations, we computed the validation cost (multinomial cross-entropy) and stopped the iterations immediately if the cost function value was increasing or decreasing by less than $1.0 \cdot 10^{-5}$. The early-stopping criteria were typically met within 20 000 iterations, corresponding to a maximum training time of 32 minutes using a normal workstation CPU only. 

\paragraph{Regularization}
We examined several regularization approaches including drop-out, $l_{1}$ and $l_{2}$ penalties on the model weights as well as the pairwise combinations of drop-out and weight penalties. A combination of drop-out regularization applied to the output layer and $l_{1}$ penalty applied to all weight variables resulted in the highest model performance and was used with all datasets.

\paragraph{Performance evaluation}
Since our main focus was on developing a model that generalizes across subjects, we used the leave-one-subject-out method to evaluate model performance. Thus, for a dataset of $m$ subjects, the training (90\% of all trials) and validation (10\% of all trials) sets comprised pooled data from randomly-selected $m-1$ subjects. The model was then applied to the data of the held-out subject, and the following two scores were computed. As all our datasets comprised an equal number of trials for each category, we used classification accuracy as the performance metric. \textit{Initial test accuracy} was defined as the proportion of correct predictions on the held-out subject. \textit{Pseudo-real-time accuracy} was defined as the mean prediction accuracy in a simulated real-time design where the model predicted new observations in batches of 20 trials and was updated after each prediction. For the true real-time BCI experiment (Experiment 3), the actual BCI accuracy (with and without model updates) is reported.

\subsection{Benchmark classifiers}
\paragraph{RBF and linear SVMs}
SVMs \citep{Vapnik2000TheTheory} are widely used in classification of MEG data \citep[e.g.][]{Gramfort2013MEGMNE-Python.,Westner2018Across-subjectsActivity}. We used incremental versions of linear and radial basis function (RBF) -kernel Support Vector Machines (SVM) as benchmark classifiers. Data preprocessing, scaling and classifier training procedure for these methods was identical to the one reported for LF-CNN and VAR-CNN. Data points from all channels/timepoints were concatenated forming a single feature vector. No additional feature extraction methods were used. We refer the reader to our previous study \citep{Halme2018Across-subjectEEG.} comparing various feature extraction methods in combination with linear classifiers for across-subject classification of MEG data. 
Nyström RBF kernel approximation was used for incremental RBF-SVM as implemented in the Scikit-Learn package\citep{Pedregosa2011}.  The SVM inverse regularization parameter $C$ and the kernel lengthscale parameter for RBF kernel $\gamma$ were set by performing a search over a 2-d grid of 5 logarithmically spaced values from $10^3$ to $10^5$ for $C$ and from $10^{-2}$ to $10^{-7}$ for $\gamma$. The classifier that gave the highest validation set accuracy was evaluated on the test set.

\paragraph{CNNs developed for EEG classification}
We used two CNN models developed for classification of EEG data. Shallow FBCSP-CNN \citep{Schirrmeister2017} is a model inspired by Filter-Bank Common Spatial Pattern (FBCSP), a state-of-the-art method for extracting band-power features in EEG/MEG. Its architecture comprises a 1-d temporal-convolution input layer (40 filters) followed by a spatial-filter layer (40 filters) and mean pooling. The outputs of the pooling layer are then combined linearly to produce label predictions by applying the softmax function. We used the Shallow FBCSP-CNN implementation provided in the Braindecode library with default parameters, only modifying temporal filters and pooling factors to match the sampling rate of our data (125 Hz). We were not able to perform pseudo-real time test for FBCSP-CNN due to the differences in implementation.

EEGnet \citep{Lawhern2018EEGNet:Interfaces} is a compact model designed specifically to optimize across-subject generalization. The model uses a combination of 1-d depth-wise and separable convolution layers (a total of 4 layers) and has been shown to generalize well across subjects in a large number of datasets. We implemented EEGNet-8 in Tensorflow following the description provided in \citet{Lawhern2018EEGNet:Interfaces} and tested it in a simulated real-time set-up, similarly to VAR-CNN and LF-CNN. 

\paragraph{Deep CNNs developed for image classification}
As an example of general-purpose deep convolutional network, we used VGG19 – a 19-layer convolutional network and winner of ImageNet Challenge 2014 \citep{Simonyan2015VeryRecognition}. The VGG19 architecture includes 5 blocks of convolutional layers followed by three fully-connected layers. Each of the 5 blocks includes a stack of several (2, 2, 4, 4, and 4) convolutional layers with (64, 128, 256, 512, and 512) 3$\times$3 convolution kernels and a 2$\times$2 max-pooling layer with stride 2. Due to the fact that the scaling of MEG data was different from that of the ImageNet dataset, we have introduced a batch-normalization layer after each block of convolutional layers to mitigate the risk of exploding or vanishing gradients. The final layer uses softmax non-linearity while all hidden layers are equipped with the ReLU non-linearity. For further details on VGG19 implementation, see \citet{Simonyan2015VeryRecognition}.

\section{Datasets}
MEG recordings were acquired using an Elekta Neuromag Vectorview (MEGIN / Elekta Oy, Helsinki, Finland) MEG system, which includes 306 sensors at 102 positions around the head; two orthogonal planar gradiometers and a magnetometer at each position. Only data from the planar gradiometers (204 channels) were used in this work. All experiments had their respective approvals from Aalto University Committee on Research Ethics.

All MEG data were sampled at 1000 Hz, band-pass filtered to 1–45 Hz and thereafter downsampled to 125 Hz, as higher bandwidth and sampling rates did not significantly improve the performance while increasing computational time.
 We used the same basic preprocessing and scaling approaches for all three experiments. Each MEG epoch was scaled independently by subtracting the mean and dividing by the standard deviation of the MEG signal in all channels during the pre-stimulus interval (the --300 ... 0-ms interval with the zero time corresponding to the stimulus onset). This approach provided reasonable scaling suitable for a real-time experiment while preserving the spatial structure of the signal.

The MEG dataset of Experiment 1 was preprocessed using a state-of-the-art off-line pipeline that is not applicable in a real-time BCI setting. External magnetic interference was suppressed and head movements compensated for using the temporally-extended signal-space separation (tSSS) method implemented in the MaxFilter software (version 2.2; MEGIN / Elekta Oy, Helsinki, Finland) \citep{Taulu2006}. Thereafter, cardiac and ocular artifacts were projected out using the FastICA algorithm as implemented in the MNE-Python software \citep{Gramfort2013}

Preprocessing the datasets from Experiments 2 and 3 was performed exclusively using methods available for real-time processing. The rationale for different preprocessing approaches was to progress from an optimal MEG pipeline (Experiment 1) to a reduced one, comprising only those methods that are available in a real-time setting (Experiment 2), and finally to demonstrate a real BCI experiment (Experiment 3). Thus, no external magnetic interference or other artifact suppression was performed in Experiments 2 and 3, and the cardiac and oculomotor artifacts, typical for MEG measurements were present in these data.

\subsection{Experiment 1: Classification of 5 types of sensory event-related fields}
Dataset 1 comprised single-trial event-related-field (ERF) responses in 7 healthy human subjects (mean age =30.1, 4 males, 3 females, 1 left-handed) to 5 types of sensory stimuli; checkerboard patterns presented in the right or left visual hemifield (Classes 1 and 2), 1-kHz 50-ms auditory tones presented to the left or right ear (pooled into Class 3), and transient transcutaneous electrical stimulation of the median nerve at the left or right wrist (Classes 4 and 5). 
This dataset comprised 500-ms segments of MEG measurements sampled at 1000 Hz (500 samples measured by 204 MEG channels) starting at the onset of each stimulus. Total trial counts per subject were $1622 \pm 322$ (mean $\pm$ SD).

\subsection{Experiment 2: Classification of event-related oscillatory activity in a 3-class motor imagery task}
Dataset 2 comprised MEG measurements of 17 healthy human subjects who performed a motor imagery task (for details, see \citet{Halme2018Across-subjectEEG.}), in which they imagined moving either their left or right hand (without any actual movement) when a visual cue was presented. The data comprised 1500-ms segments of MEG measurements sampled at 1000 Hz (1500 samples measured by 204 MEG channels) starting at the onset of the visual cue. Based on the measured MEG signal, we decoded whether the subject imagined moving his/her left or right hand, or nothing at all (rest condition). Total trial counts per subject were $120 \pm 5$ (mean $\pm$ SD).

\subsection{Experiment 3: Real-time motor imagery BCI}

Here we applied the same experimental paradigm as in Experiment 2, but with true real-time decoding and updating of the model. For this experiment, a network trained using a subset of trials from 17 subjects (Experiment 2) including only two classes (left vs. right motor imagery) was integrated into a real-time motor imagery BCI. Two subjects performed a task where they had to imagine moving the left or right hand following the presentation of a visual cue (an arrow pointing to the left or to the right). The VAR-CNN model which showed the highest performance in Experiment 2 performed 2-class classification (left vs. right hand motor imagery) in real time.

None of the subjects had used motor imagery-based BCIs before or were involved in the collection of the MEG data used to train the model, i.e in Experiment 2. 
The experiment comprised three sessions. In the first session, the classifier was applied without the online updates to estimate the baseline performance. In sessions 2 and 3, the model parameters were updated following each trial where the classifier decoded the subject's intention correctly. The update was performed with a single back-propagation step using the MEG data from this trial and the associated label. Since our subjects had no prior experience with motor imagery tasks we chose to avoid updating the classifier after trials where the misclassification was due to the subject struggling to perform the mental task (as opposed to classifier errors). Updating the decoder with these erroneous trials would likely yield decreased decoding accuracy in subsequent trials. For the same reason we constrained this experiment to have only two classes. Trials in each session were presented in three different pre-defined sequences (50 trials per session) such that the true labels were available for real-time accuracy estimation and incremental model optimization.

\section{Results}

\subsection{Experiment 1}
In this 5-class decoding of sensory ERFs, VAR-CNN outperformed other models in terms of accuracy on a pooled validation set ($95.8\% \pm 0.7\%$), when applied to held-out subjects ($85.9\% \pm 7.4\%$), and in pseudo-real time tests ($94.2\% \pm 2.8\%$). The results from LF-CNN were the second best, and RBF-SVM was the closest benchmark. Detailed statistical comparisons between the performance of LF-CNN, VAR-CNN and the benchmark classifiers in Experiment 1 are summarized in Supplementary Table 2.

In the simulated real-time experiments, the CNN-based models EEGNet-8, LF-CNN and VAR-CNN significantly improved their performance ($+12.4$,$+10.2$, and $+8.3$ percent points, respectively compared to the initial test accuracies. In case of SVM classifiers, this improvement was considerably smaller ($+6.8 $ for Linear SVM, and $+1.2$ percent points for RBF-SVM). These results are summarized in Table \ref{tab:exp1_5class}.

\begin{table*}[h!]
\centering

\caption{Across-subject performance in a 5-class sensory stimulation task. Grand-average accuracy scores (mean $\pm$ SD) from leave-one-subject-out cross-validation. Highest-performing model in each test is indicated in bold.}

\begin{tabular}{lrrrr}
\toprule
  Model & Validation (\%) & Initial test (\%) & Pseudo-real-time (\%) \\
  \midrule
LF-CNN & $95.0 \pm 0.8$ & $83.1 \pm 8.4$ & $93.3 \pm 3.6$ \\
VAR-CNN & $\mathbf{95.8 \pm 0.7}$ & $\mathbf{85.9 \pm 7.4}$& $\mathbf{94.2 \pm 2.8}$\\
Linear SVM &$93.3 \pm 1.2$& $80.2 \pm 9.7 $ &$87.0 \pm5.4 $\\
RBF-SVM & $93.6 \pm 1.7$ & $82.7 \pm 8.3$ & $83.9 \pm 8.4$ \\
ShallowFBCSP-CNN\ & $85.3 \pm 2.4$& $60.1 \pm 11.7$ & n.a. \\
EEGNet-8 & $88.7 \pm 2.0$ & $76.8 \pm 11.7$ & $89.2 \pm5.0$\\
VGG19 & $80.5 \pm3.3$& $70.1 \pm 12.8$&$73.9 \pm 10.5$\\
 \bottomrule
\end{tabular}
\label{tab:exp1_5class}
\end{table*}

\subsection{Experiment 2}
With the 3-class motor imagery dataset, VAR-CNN again gave the highest classification accuracies when applied to pooled validation set ($86.7\% \pm 7.4\%$), held-out subjects ($76.3\% \pm 6.8\%$) and pseudo real-time tests ($82.3\%\pm 6.1\%$). The closest bechmarks classifiers were RBF-SVM for held-out subjects ($74.1\% \pm 8.4\%$), and EEGnet-8 for the validation set ($80.8\% \pm 2.4\%$) and pseudo real-time test ($80.9\% \pm 6.7\%$). Detailed statistical comparisons between the performance of LF-CNN, VAR-CNN and the benchmark classifiers in Experiment 2 are summarized in Supplementary Table 3. 
Similarly to Experiment 1, EEGNet-8, LF-CNN and VAR-CNN were able to significantly improve their performance in a simulated real-time test using incremental updates ($+8.9$,$+3.8$, and $+5.7$ percent points, respectively) compared to the initial test accuracies. The linear SVM classifier performance improved to a smaller extent through pseudo-real-time updates ($+3.2 $ percent points), while the performance of RBF-SVM did not improve at all ($-0.5$ percent points). 

\begin{table*}[h!]
\centering

\caption{Across-subject performance in a 3-class motor imagery task. Grand-average accuracy scores (mean $\pm$ SD) from leave-one-subject-out cross-validation. Highest-performing model in each test is indicated in bold.}

\begin{tabular}{lrrrr}
\toprule
 Model & Validation (\%) & Initial test (\%) & Pseudo-real-time (\%) \\
 \midrule
 LF-CNN & $84.3 \pm 2.7$& $74.2 \pm 6.5$& $78.0\pm 6.5$ \\
VAR-CNN & $\mathbf{86.7 \pm 7.4}$ & $\mathbf{76.3 \pm 6.8}$& $\mathbf{82.3 \pm 6.1}$\\
Linear SVM &$76.9 \pm 3.0$&$ 68.2 \pm 7.2$&	$71.4\pm7.3	$\\
RBF-SVM & $80.3 \pm 2.6$ & $74.1 \pm 8.4$ & $73.6 \pm 8,8$\\
ShallowFBCSP-CNN & $70.2 \pm 4.1$ & $60.2 \pm 10.3$ & n.a. \\
EEGNet-8 & $80.8 \pm 2.4$ & $72.1 \pm 5.8$ & $80.9 \pm6.7$\\
VGG19 &$71.4 \pm 9.6 $& $60.2 \pm 6.8$& $57.4 \pm9.2$\\

\bottomrule
\end{tabular}

\label{tab:exp2}

\end{table*}

\subsection{Experiment 3: Real-time motor imagery BCI}
Results of the real-time application of VAR-CNN are summarized in Table \ref{tab:exp3}. Comparing the accuracies achieved by VAR-CNN with and without stochastic updates clearly shows the significant increase in performance. We note that since the subjects had no prior experience with motor imagery BCIs, the improvement in performance after Session 1 may be partly attributed to their improved motor imagery skills.
\begin{table*}[h!]
\centering

\caption{VAR-CNN classification accuracy in the real-time motor imagery BCI experiment.}

\begin{tabular}{lccc}
\toprule
 Subject & Run 1, no updates (\%) & Run 2, online updates (\%) & Run 3, online updates (\%)\\\midrule
s01 & $80.0$ & $88.0$ & $92.0$  \\
s02 & $62.0$ & $90.0$ & $ 82.0$   \\
\bottomrule
\end{tabular}

\label{tab:exp3}
\end{table*}

\subsection{Interpretation of learned model parameters}
\begin{figure*}
\centering
\includegraphics[width=0.9\linewidth]{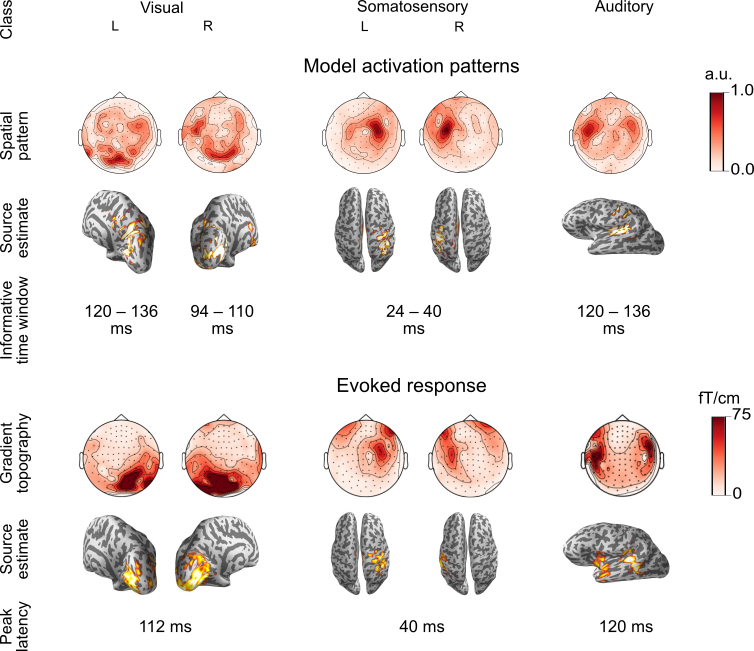}
\caption{Interpretation of informative LF-CNN model parameters in a representative subject from Experiment 1. \textbf{A.} Components having the maximum contribution to the decoding of each class were extracted from the model and interpreted in terms of their spatial topographies (top), source estimates (middle) and latency estimates (bottom).
\textbf{B.} Spatial topographies(top), source estimates (middle) and peak latencies (bottom) of the corresponding evoked responses. Source estimates visualization is thresholded to 95\% of the peak source activity.}
\label{fig:interpretability}
\end{figure*}

Figure \ref{fig:interpretability} shows the activation patterns and the corresponding informative time-windows of the components with the maximum contribution to the decoding of each class in the LF-CNN model, trained on the pooled data from Experiment 1 and updated using the pseudo-real time update procedure described above on single held-out subject. For all of the five classes, our model extracted spatial patterns whose source estimates showed overall good correspondence to the locations and lateralizations of the peaks of the evoked responses (Figure \ref{fig:interpretability}).

Similarly, activation patterns extracted from the data of Experiment 2 following a similar training and updating procedure resulted in spatial patterns focused over parietal as well as the more posterior occipital sensors contralateral to the imagined movement. Interestingly, spectral estimates of the most informative latent components resulted in density estimates peaking at around 10 Hz corresponding to well-known $\mu$-rhythm desynchronization associated with motor imagery  (Figure \ref{fig:spectra}). Although anatomical information was not available for this dataset these results suggest that apart from motor and pre-motor cortices more posterior sources might also contribute to motor imagery. 

\begin{figure*}
\centering
\includegraphics[width=1\linewidth]{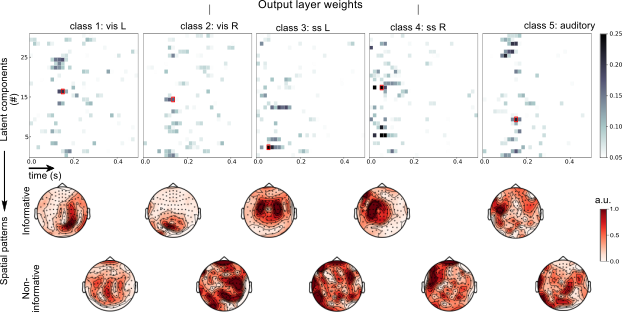}
\caption{Identification of informative latent components in a single representative subject from Experiment 1. The weights of the contributions of the final layer to each class are represented as a raster plot (top) with rows corresponding to index of latent component and columns corresponding to (pooled) time points. Informative components (middle) are found by identifying single features (indicated by red boxes) with maximum positive weight for each class. Non-informative components are defined as having minimal absolute sum of weights across all classes. Component topographies are scaled to interval [0,1] for comparability.}
\label{fig:non_informative}
\end{figure*}

We also estimated spatial properties of the latent components that had the least overall contribution to any of the classes. Inspecting the weights of the output (classification) layer (Figure \ref{fig:non_informative}) further allowed us to identify 5 components which provided minimum contribution to either class. Although none of these components directly corresponded to known signatures of e.g. oculomotor artifacts, their overall limited contribution to either class may suggest that these patterns were used for out-projecting irrelevant activity.

\section{Discussion}

In this paper, we report two neural network models optimized for across-subject classification of electromagnetic brain signals. 
These models outperform traditional approaches as well as more complex deep neural networks in terms of accuracy, across-subject generalization, and simulated real-time performance. Arising from inter-individual variability in structural and functional cortical anatomy, across-subject generalization has proved to be a challenging problem for machine-learning methods applied to EEG and particularly to MEG data due to its higher spatial resolution. To illustrate the severity of this problem, we estimated the correlations between the spatial patterns extracted from our model fine-tuned to individual subjects. Mean spatial correlation coefficient between the most consistent components was \textit{r} = 0.67 in Experiment 1 ($N=7$) and \textit{r} = 0.52 in Experiment 2 ($N=17$).
One advantage of our models is that they introduce reasonable assumptions based on the signal generation model. These assumptions include e.g. linear separability in spatial domain, consistency of temporal or spectral properties of the signal across trials etc. Thus, we would expect it to be most sensitive to features common across subjects while allowing for reasonable variability. 

\begin{figure}
\centering
\includegraphics[width=0.85\linewidth]{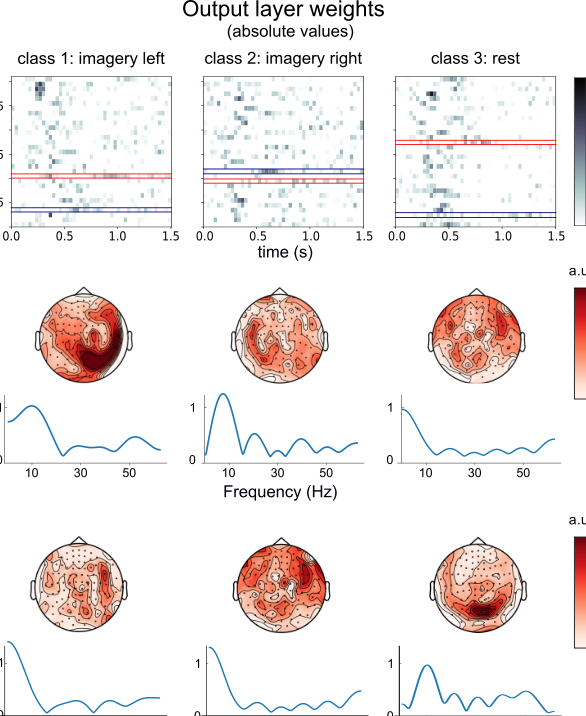}
\caption{Interpretation of informative LF-CNN model parameters in a representative subject from Experiment 2. Latent components having the maximum positive (red) and negative (blue) sum of weights over all time points for each class were extracted from the model and interpreted in terms of their spatial topographies, and spectral estimates.}
\label{fig:spectra}
\end{figure}

We further show that the relatively low complexity of our LF-CNN architecture allows interpretation of the learned model parameters in terms of the underlying neural activity. Such interpretation can prove to be a convenient tool for quickly exploring complex, high-dimensional MEG and EEG datasets and ultimately allow extracting more information from these rich data. Interpretation of discriminative models can provide valuable insights into these data by e.g. allowing to dissociate "most active" sources from those that contribute the most to discrimination. One potential application of these methods could be investigating neural sources contributing to BCI control. This argument, however, holds only for LF-CNN, because the more complex VAR-CNN variant allows to capture non-linear interactions between the latent sources and thus cannot be interpreted in a straightforward way.

Moreover, we propose a procedure where the classifier is initialized from the data of other subjects and then updated online during the real-time experiment. We demonstrate that using this approach our models perform accurately on new subjects in a real-time BCI experiment, allowing new subjects to efficiently use the system without separate calibration. Importantly, this procedure allows to omit a dedicated BCI calibration session, facilitating the use of BCIs in research and clinical settings. 

The results reported here show considerable improvement in performance compared to our previous results in across-subject decoding of MEG. Using state-of-the-art feature extraction methods in combination with linear classifiers \citet{Halme2018Across-subjectEEG.} achieved a classification accuracy of 70.6\%. Our best-performing model was able to classify 3 classes of stimuli on the same dataset with 76.3\% accuracy. Furthermore, adding incremental updates to the classifier allowed us to achieve even higher accuracy (82.3\%) in a simulated real-time test. We confirmed that the feasibility of the latter approach in a real-time BCI experiment, achieving similarly high accuracy in Experiment 3.

Several studies report applying deep neural networks to single-trial classification of non-invasive neurophysiological measurements, typically multichannel EEG data. Most successful models make use of the spatiotemporal structure of EEG \citep{Bashivan2015LearningNetworks,Hajinoroozi2016,Lawhern2018EEGNet:Interfaces,Schirrmeister2017} including the real-time applications \citep{Burget2017ActingSkills,Fahimi2019Inter-subjectBCI}. \citet{Bashivan2015LearningNetworks} exploit the spatiotemporal and spectral structure of EEG by transforming the signals into a sequence of multidimensional spatio–spectral images using time–frequency and polar transforms to achieve significant performance improvement in classifying cognitive load. By contrast, we use a simpler linear projection in the spatial domain. Similarly to our approach, \citet{Hajinoroozi2016} tested spatial ICA as a preprocessing step followed by temporal 1-d convolution filters. However, such a combination did not result in a significant improvement in model performance in that study, as compared to using EEG channel data as the input. We argue that -- when performed separately from the classification -- ICA decomposition may not be optimal due to the independence assumption which may not hold for real EEG and MEG signals. In our design, analogous linear projection in the spatial domain was obtained by back-propagation. When trained in conjunction with the frequency filters in the temporal convolution layer, such projection results in a separable decomposition related to a combination of Linear Discriminant Analysis \citep{McLachlan1992DiscriminantRecognition}, and Spatio--Spectral Decomposition \citep{Nikulin2011} used in EEG--MEG analysis.

\citet{Lawhern2018EEGNet:Interfaces} introduced the EEGNet model as a compact, interpretable CNN architecture that can be applied to different EEG-based BCI paradigms (including both evoked and induced responses). EEGNet has also been shown to generalize well to held-out subjects. Importantly, in our simulated real-time tests, EEGNet came close to the accuracies of our models, demonstrating the advantages of neural networks in combination with stochastic optimization.

Similarly to other models \citep[e.g.][]{Schirrmeister2017}, \citet{Lawhern2018EEGNet:Interfaces} applied 1-d temporal convolutions to the raw EEG channel data. Our results also indicate that models using 1-d convolutions are better suited for classification of MEG data than the deeper image classification networks relying on 2-d convolutions. In contrast to previous studies, we apply spatial decomposition first, followed by a temporal depth-wise (LF-CNN) or spatio–temporal (VAR-CNN) convolution. The motivation for this is three-fold: first, it allows effective spatial decorrelation and dimensionality reduction; second, when nested into the CNN architecture, it allows the network to learn and project out physiological artifacts similarly to other linear projection methods; finally, it can contribute to improved generalization across subjects by increasing model robustness to inter-individual differences in spatial distributions of the informative features. The latter argument is particularly important for MEG, since its spatial resolution is considerably higher than that of EEG, and even minor differences in source location or orientation may change the signal patterns at the sensors and lead to different channels being most sensitive to the brain activity of interest. This is the likely reason why the CNN models optimized for EEG decoding did not perform optimally in our study.

Our results demonstrate that with an adequate choice of hyper-parameters, support vector machines may perform equally well as the best-performing CNN models in off-line classification. We thus recommend using SVMs as benchmark models in future studies. Our simulated real-time tests, however, demonstrate that incremental versions of SVMs do not gain as much in performance from the real-time updates as the CNN-based models do. This result probably reflects the fact that incremental SVMs rely on kernel approximations and thus their fine-tuning to individual subjects is limited.

Thus, the proposed models outperformed the benchmark methods and provided means to investigate spatial and temporal patterns that contribute to discrimination between the stimuli. Inspecting these patterns may prove a useful tool to obtain fast preliminary estimate of the neural activity informing the classification.  Yet, the degree of correspondence between the patterns informing the classification and the actual neural activity is not clear and requires  further systematic testing. 

In this study, we restrict the interpretation of the model parameters to the single most-informative latent component. However, it is expected that several distinct components can contribute to a single class and provide valuable information (as e.g. shown in Figure \ref{fig:spectra}). We suggest two heuristic approaches to identify such informative components. Similarly, we limit ourselves to the interpretation of the simpler variant (LF-CNN) of the proposed networks due to its low complexity and predominantly linear activation functions. A more comprehensive model investigation would require a more systematic approach, suggested e.g. in \citet{Kindermans2018LearningPaternAttribution} and \citet{Alber2018INNvestigateNetworks}.

Finally, we show that the proposed models outperform other neural networks designed for EEG classification when applied to MEG data. Given a large number of channels and a higher spatial resolution MEG datasets may benefit from these designs. It remains to be seen, however, how the proposed models can perform when applied to EEG data. 

\section{Conclusions}
We have introduced a Convolutional Neural Network model optimized for off- and on-line (real-time) classification of MEG data. Incorporating prior knowledge about the processes generating MEG observations allowed us to substantially reduce model complexity while preserving high accuracy and interpretability. We show that this model successfully classifies evoked as well as oscillatory activity and generalizes efficiently across subjects. When combined with incremental real-time model updates, the time-consuming calibration sessions in MEG-based brain--computer interfaces could be omitted provided that a sufficient amount of training data from other subjects is available.

\section{Acknowledgements}
Research reported here was supported by the Academy of Finland, Grant/Award Number: “NeuroFeed” / 295075 and the European Research Council under ERC Grant Agreement n. 678578. The content is solely the responsibility of the authors and does not necessarily represent the official views of the funding organizations.

\section*{References}


\end{document}